\documentclass[11pt]{article}

\usepackage[]{acl}

\usepackage{times}
\usepackage{latexsym}

\usepackage[T1]{fontenc}

\usepackage[utf8]{inputenc}

\usepackage{microtype}

\usepackage{inconsolata}

\usepackage{graphicx}
\usepackage{booktabs}
\usepackage{multirow}
\usepackage{graphicx}
\usepackage{xcolor}
\usepackage{colortbl} 

\usepackage{booktabs}
\usepackage{multirow}
\usepackage[table]{xcolor} 
\usepackage{graphicx}
\definecolor{ourbg}{RGB}{236, 245, 255}     
\definecolor{headergray}{RGB}{245, 247, 249} 
\definecolor{improvecolor}{RGB}{46, 139, 87} 
\definecolor{textgray}{RGB}{60, 60, 60}      

\usepackage{algorithm}
\usepackage{algpseudocode}
\usepackage{booktabs,makecell,enumitem}
\usepackage{array}      
   \usepackage{booktabs}    
\usepackage{tabularx}   
\usepackage{array}       
\usepackage{caption}
\usepackage{booktabs}
\usepackage{threeparttable}
\usepackage{graphicx}    
\usepackage{multirow}    
\usepackage{adjustbox}   
\usepackage{colortbl}  
\usepackage{subcaption}  
\usepackage{booktabs}    
\usepackage{caption}     
\usepackage{placeins} 

\usepackage{booktabs}
\usepackage{amsmath}
\usepackage{algorithm}
\usepackage{algpseudocode}
\usepackage[table]{xcolor}
\usepackage[utf8]{inputenc}
\usepackage{listings}
\usepackage{caption}

\definecolor{codegreen}{rgb}{0,0.6,0}
\definecolor{codegray}{rgb}{0.5,0.5,0.5}
\definecolor{codepurple}{rgb}{0.58,0,0.82}
\definecolor{backcolour}{rgb}{0.95,0.95,0.92}
\definecolor{codeblue}{rgb}{0.25,0.5,0.7}
\definecolor{codered}{rgb}{0.8,0.1,0.1}

\newcommand{\ourmethod}{VideoPro }

\usepackage{textcomp}
\makeatletter
\DeclareFontFamily{OMS}{zi4}{}
\DeclareFontShape{OMS}{zi4}{m}{n}{<->ssub*cmsy/m/n}{}
\DeclareFontShape{OMS}{zi4}{b}{n}{<->ssub*cmsy/b/n}{}
\DeclareFontShape{OMS}{zi4}{bx}{n}{<->ssub*cmsy/b/n}{}
\makeatother

\usepackage{booktabs, multirow, array, makecell, adjustbox, colortbl, xcolor}
\usepackage{graphicx, subcaption, float, wrapfig}
\usepackage{pifont}
\usepackage{fixmath, mathtools, nicefrac, amsfonts}
\usepackage{longtable, comment, arydshln}
\usepackage{caption}
\usepackage[capitalize]{cleveref}
\usepackage{listings}
\usepackage[most]{tcolorbox}
\usepackage{tabularx}
\usepackage{marvosym}
\usepackage{placeins}

\definecolor{lightblue}{HTML}{DAEFF9}
\definecolor{bisque}{rgb}{1.0, 0.89, 0.77}
\definecolor{ForestGreen}{rgb}{0, 0.69, 0.31}
\definecolor{NavyBlue}{rgb}{0, 0.44, 0.75}

\newlength\savewidth

\lstdefinestyle{prompt}{
    basicstyle=\ttfamily\fontsize{7pt}{8pt}\selectfont,
    frame=none,
    breaklines=true,
    backgroundcolor=\color{lightgray},
    breakatwhitespace=true,
    breakindent=0pt,
    escapeinside={(*@}{@*)},
    numbers=none,
    numbersep=5pt,
    xleftmargin=5pt,
}

\tcbset{
  aibox/.style={
    top=10pt,
    colback=white,
    colframe=black,
    colbacktitle=black,
    enhanced,
    center,
    attach boxed title to top left={yshift=-0.1in,xshift=0.15in},
    boxed title style={boxrule=0pt,colframe=white,},
  }
}
\newtcolorbox{AIbox}[2][]{aibox, title=#2,#1}

\title{VideoPro: Adaptive Program Reasoning for Long Video Understanding}

\author{Chenglin Li\textsuperscript{1,4}, Feng Han\textsuperscript{2,4}, Yikun Wang\textsuperscript{2,4}, Ruilin Li\textsuperscript{3,5}, Shuai Dong\textsuperscript{4}, \\
Haowen Hou\textsuperscript{4}, Haitao Li\textsuperscript{1,5}, Qianglong Chen\textsuperscript{1}, FengTao\textsuperscript{4}, Jingqi Tong\textsuperscript{2,4}, \\
Yin Zhang\textsuperscript{1,4}, Jiaqi Wang\textsuperscript{4} \\
  \textsuperscript{1}Zhejiang University, \textsuperscript{2}Fudan University, \textsuperscript{3}Wuhan University, \\
\textsuperscript{4}Shanghai Innovation Institute \\
}

\begin{document}
\maketitle


\begin{abstract}

Understanding long videos remains challenging due to the sparsity of visual evidence relevant to a given query. Prior work has explored program-based visual grounding, typically relying on executable programs generated by auxiliary large language models. However, when scaling to long videos, existing approaches face several critical limitations: (1) frame-centric vision modules are often insufficient for long video processing; (2) naively applying program-based reasoning to all queries incurs considerable computational overhead; and (3) errors arising from low-confidence predictions and imperfect program execution are difficult to recover from.
To address these challenges, we propose VideoPro, a unified framework that enables VideoLLMs to adaptively reason over long videos and refine their predictions through executable programs. VideoPro first performs adaptive reasoning, dynamically determining whether a query can be resolved directly by the native VideoLLM or requires explicit multi-step program reasoning. For complex queries, the model decomposes the task into executable programs that invoke specialized vision modules for precise temporal and semantic grounding. To further improve robustness, VideoPro incorporates a self-refinement mechanism that leverages execution feedback and confidence signals to correct erroneous executions and refine low-confidence reasoning programs. By tightly integrating adaptive reasoning with self-refinement, VideoPro consistently outperforms prior methods across multiple long-video understanding benchmarks, yielding an average 6.7\% improvement for Qwen3-VL-8B.
\end{abstract}

\begin{figure*}[t]
    \centering
     \includegraphics[width=1.0\linewidth]{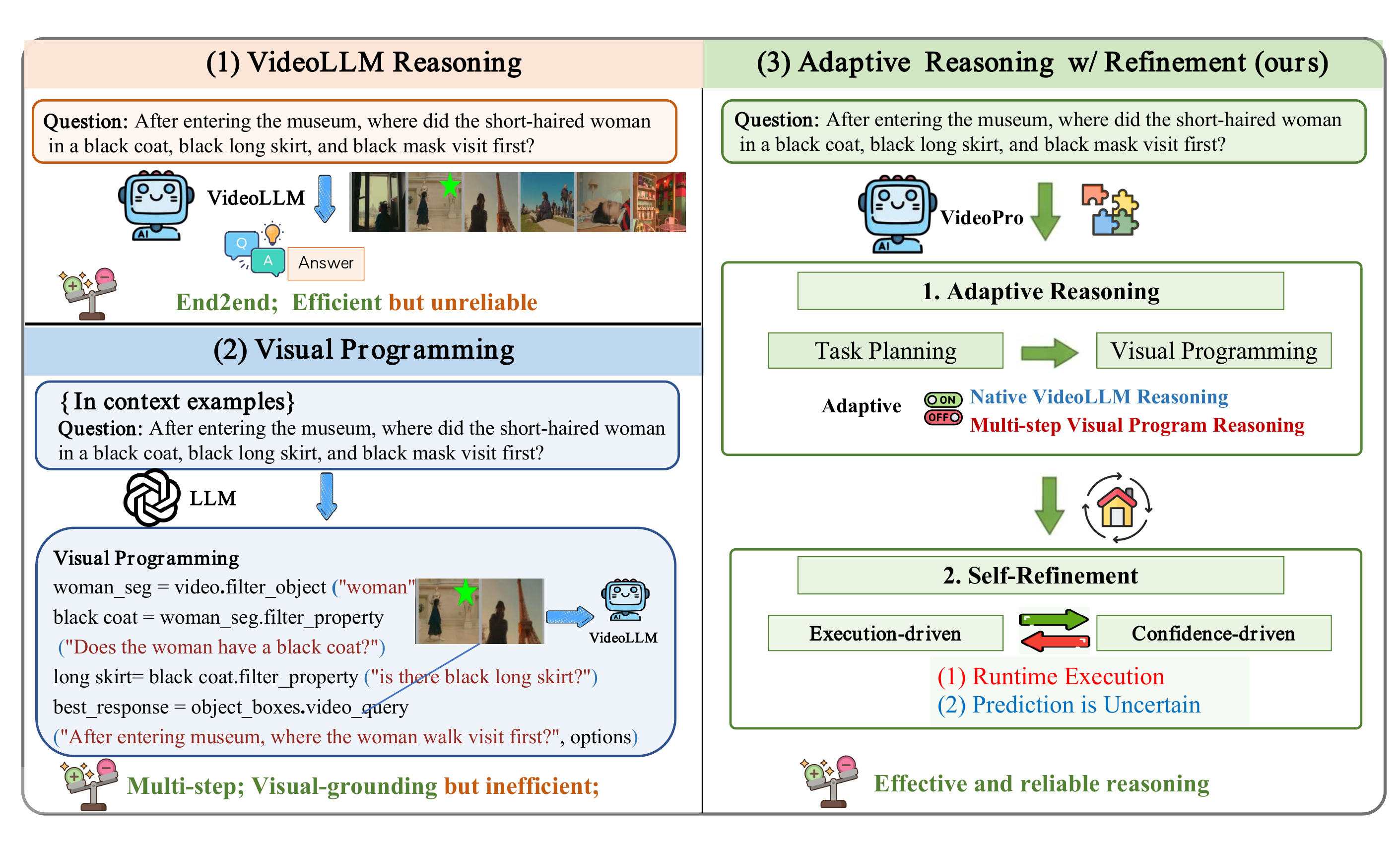}
\caption{Comparison of prior methods and VideoPro: effective and reliable adaptive reasoning with refinement.}

    \label{figure:compare}
\end{figure*}

\section{Introduction}

Long-video understanding is fundamentally challenged by the sparsity and long-range dispersion of query-relevant visual evidence. Effective reasoning, therefore, requires precise temporal grounding and multi-step integration across distant segments, rendering dense frame-level processing computationally intractable. Program-based visual grounding has been proposed to support explicit multi-step reasoning by executing LLM-generated programs over modular vision components~\citep{gupta2023visual,suris2023vipergpt,mahmood2024vurf,gao2024fine} (Figure~\ref{figure:compare}). However, when scaled to long videos, existing methods exhibit three critical limitations.
(1) Frame-centric vision modules are insufficient for long-range temporal and semantic grounding.
(2) Uniform program invocation is inefficient: many queries can be answered directly by native VideoLLMs, which already achieve nearly 90\% accuracy in high-confidence cases without program execution (Figure~\ref{fig:confidence_stats}).
(3) Error recovery is fragile: low-confidence predictions and imperfect executions can cascade under static, one-shot program pipelines, with limited mechanisms for correction.

To address these challenges, we propose VideoPro, a unified framework that synergizes adaptive reasoning with refinement for efficient long-form video understanding. \ourmethod functions as a dynamic planner, determining on-the-fly whether a query can be resolved by the native VideoLLMs or requires multi-step visual programming. For complex queries, it decomposes the task into structured sub-tasks and selectively invokes vision modules to retrieve, localize, and inspect relevant information. To support the latter, \ourmethod integrates a library of general video modules, including multimodal retrieval, temporal localization, and fine-grained visual extraction. To further improve robustness, \ourmethod incorporates a self-refinement mechanism, which revises failed executions and low-confidence reasoning programs, ensuring robust reasoning against static programs.

To instantiate this framework, we leverage advanced LLMs to synthesize high-quality visual programs and use the native VideoLLM's initial prediction to decide whether a query can be solved directly or should be routed to explicit visual program reasoning. We categorize execution programs into failures, successful executions with incorrect predictions, and correct predictions to construct a \emph{reason-and-refine} supervision dataset. We then train a unified VideoLLM to internalize the ability of adaptive reasoning and self-refinement, further optimized using Group Relative Policy Optimization (GRPO). Extensive experiments demonstrate that \ourmethod achieves superior stability and performance on long-form video benchmarks, surpassing GPT-4o on LVBench with 49.7\% accuracy and boosting Qwen3-VL-8B's performance by 6.7\% on average. Our main contributions are:
\begin{itemize}
    \item Effective Adaptive Reasoning. We introduce a query-level adaptive mechanism that dynamically selects between native VideoLLM reasoning and multi-step visual program reasoning, achieving an optimal balance between performance and efficiency.
    \item Reliable Refinement. We propose a self-refinement mechanism guided by execution and confidence signals, enhancing the reliability of program-based reasoning.
    \item Promising Performance on Long-Form Video. We design a suite of general video modules for retrieval, localization, and inspection, enabling our \ourmethod to achieve consistent improvements across multiple challenging long-video benchmarks.
\end{itemize}

\section{Related Work}
\label{sec:related_work}

\subsection{Long-form Video Understanding}
Understanding long-form videos requires identifying and connecting sparse evidence distributed across extensive temporal horizons~\citep{wu2024longvideobench,fu2024video}. 
Current Video-LLMs typically extend image-based multimodal frameworks~\cite{liu2023visual} by sampling multiple frames to capture temporal dynamics~\cite{li2023videochat,zhang2023video,lin2023video,li2024llava,song2023moviechat}. 
These models generally align visual features with the language space using a frame-wise encoder followed by a projection module. 
Despite recent advancements in spatio-temporal representations~\citep{li2023videochat,bai2025qwen2}, dense processing of long videos remains computationally prohibitive. 
To mitigate this, alternative strategies leverage textual summaries or keyframe captioning~\citep{zhang2023simple,wang2024videoagent,wang2025videotree}; however, while these methods enhance scalability, they often sacrifice fine-grained visual details and involve complex, multi-step inference. To address this, some strategies use captioning or keyframe summarization to create textual representations for LLMs~\citep{zhang2023simple,wang2024videoagent,wang2025videotree}, which improve scalability but can lose fine-grained temporal details and require multiple inference steps. \ourmethod addresses this trade-off by leveraging native VideoLLMs for video understanding while selectively invoking targeted modules to capture key visual evidence when necessary.

\subsection{Visual Program Reasoning}
Visual program reasoning empowers LLMs to decompose queries into executable programs that orchestrate perception tools~\citep{gupta2023visual,choudhury2023zero}. However, existing frameworks are predominantly tailored for images or short clips and often suffer from brittleness in complex environments. For instance, ViperGPT~\citep{suris2023vipergpt} integrates visual modules for image and short-video QA~\citep{choudhury2023zero}, while VURF~\citep{mahmood2024vurf} enhances program reliability. These efforts position LLMs as general-purpose planners for decomposing complex tasks into interpretable steps. We extend this paradigm by introducing Adaptive Reasoning to determine when to invoke programs for long-form videos dynamically.

\subsection{Adaptive Reasoning and Refinement}
Adaptive computation allocates resources based on problem complexity, akin to the ``System 1 vs. System 2'' duality~\citep{evans2003two,xiao2025fast,sun2025fast,zhang2025othink,sun2024visual}. Complementary to this, self-refinement enhances reliability by revising outputs based on feedback~\citep{madaan2023self}. While prior works often treat routing and refinement as separate prompting strategies, \ourmethod unifies them into a learnable framework. We train a unified VideoLLM to jointly perform Adaptive Reasoning and Iterative Refinement, ensuring robust performance for long-form video understanding.
\begin{figure}[t]
    \centering
    \includegraphics[width=1.0\linewidth]{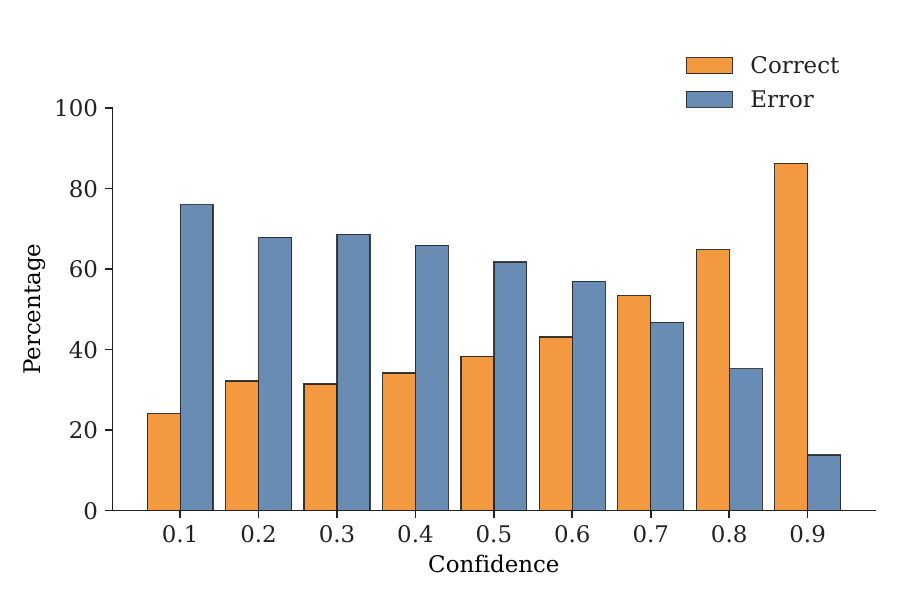}
    \caption{Distribution of correct vs.\ error predictions across confidence on LongVideoBench. The proportion of correct predictions exceeds errors in \([0.7, 0.8)\) interval, and exceeds \(90\%\) when confidence is above \(0.9\).}
    \label{fig:confidence_stats}
\end{figure}

\section{Method}
\subsection{Overview}
\label{sec:method_overview}

\begin{figure*}[t]
\centering
\includegraphics[width=1.0\textwidth]{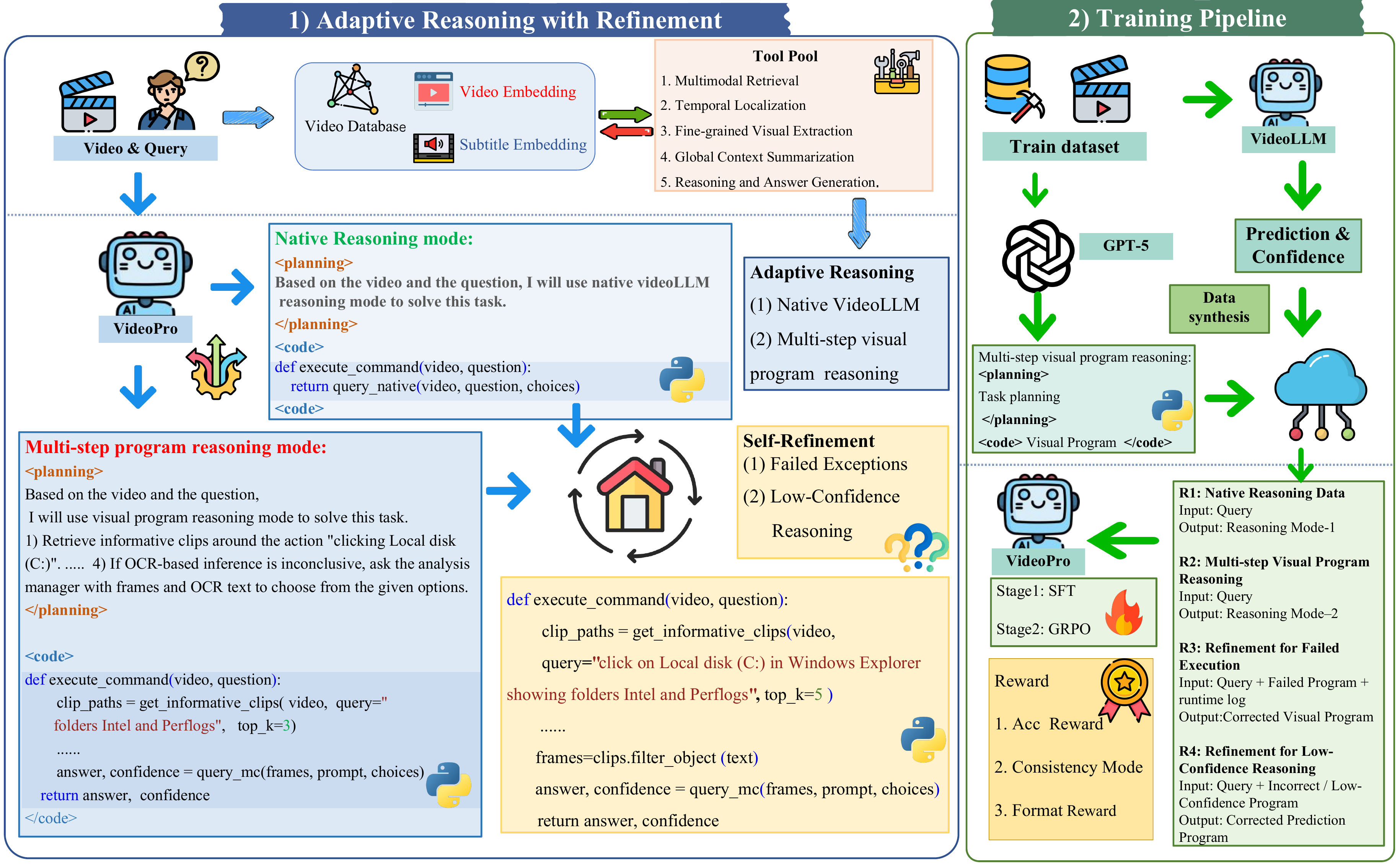} 
\caption{\textbf{(a) Adaptive Reasoning \& Self-Refinement:} VideoPro dynamically selects between Native VideoLLM and Multi-step visual program reasoning based on query complexity. Self-refinement is employed to correct failed executions and low-confidence reasoning programs.
\textbf{(b) Training Pipeline:} The process involves (1) SFT on the \textit{reason-and-refine} dataset, and (2) GRPO to optimize rewards for correctness, format validity, and consistency.}
\label{fig:overview}
\end{figure*}

Given a long-form video $V=\{f_1, \ldots, f_T\}$ and a natural language query $Q$, our goal is to generate an accurate answer $A$ with low computational cost. In long-form videos, task-relevant evidence is often sparse and temporally dispersed, making dense frame-wise processing inefficient and largely redundant. We thus cast long-form video understanding as a cost-aware dynamic inference problem centered around two core mechanisms: \textit{adaptation} and \textit{refinement}. As illustrated in Figure~\ref{fig:overview}, \ourmethod implements a unified adaptive reason-refine process with two coupled components. 
(1) \textbf{Adaptive Reasoning:} VideoPro adaptively routes each query to either \emph{native reasoning} (direct answer generation with native VideoLLM reasoning) when it is confident, or \emph{program reasoning} that explicitly composes and executes multi-step video modules when tool use is necessary. 
(2) \textbf{Self-Refinement:} conditioned on runtime feedback (e.g., execution failures) and low-confidence predictions, VideoPro revises the generated program and re-executes it to recover from errors and improve answer reliability.

\subsection{Video Module Library}
\label{sec:module_library}
Prior approaches are mostly frame-centric, often running VQA models (e.g., BLIP-2) on every single frame to find key visual information (e.g., checking for the presence of a text object)~\cite{choudhury2023zero,suris2023vipergpt}. While this works for short clips, it is extremely slow and inefficient for long videos. In contrast, \ourmethod adopts a general video module library tailored for long contexts. Rather than checking every frame blindly, we adopt a coarse-to-fine pipeline: progressing from global semantic retrieval to precise temporal localization, and finally to fine-grained visual information. We structure the capabilities of \ourmethod into five core modules: Multimodal Retrieval, Temporal Localization, Fine-grained Visual Extraction, Global Context Summarization, and Reasoning and Answer Generation. In addition to these core modules, \ourmethod utilizes basic Python operations as underlying utilities. More details for each vision module are provided in Appendix~\ref{appendix:modules}.

\subsection{Visual Programming}
\label{sec:modules}
To facilitate multi-step program reasoning with vision modules, \ourmethod frames long-video understanding as the execution of a synthesized program $P$ over a module library $\mathcal{M}$. We formally define the execution interface as:
\begin{equation}
(\hat{A}, c) = \text{Exec}(P \mid V, Q, \mathcal{M}),
\end{equation}
where $\hat{A}$ denotes the predicted answer, $c \in [0,1]$ is the associated confidence score. This explicit formalism uncovers intermediate reasoning steps, providing essential diagnostic signals for the subsequent refinement stage. Specifically, the confidence score $c$ is derived from the distribution as:
\begin{equation}
c = \exp \left( \frac{1}{L} \sum_{i=1}^{L} \log p_\theta(\hat{a}_i \mid V, Q, \hat{a}_{<i}) \right),
\label{eq:confidence}
\end{equation}
where $\hat{A}=\{\hat{a}_1, \dots, \hat{a}_L\}$ denotes the generated response of length $L$.

\subsection{Adaptive Reasoning}
\label{sec:reason_stage}

To balance reasoning effectiveness with computational efficiency, \ourmethod introduces Adaptive Reasoning via query-conditioned task planning. Given a query $Q$, the model acts as a planner that adaptively selects one of two reasoning modes:

\begin{itemize}[leftmargin=*,noitemsep]
   \item \textbf{Native VideoLLM reasoning:} If the VideoPro determines that the query can be answered using holistic understanding with native VideoLLM, it will generate programs that only use the query\_native module as a single-step program call. Without multi-step reasoning, the VideoLLM will generate the prediction directly from the video frames, avoiding unnecessary vision module invocations.
    \item \textbf{Multi-step visual program reasoning:} If VideoPro identifies a need for multi-step and video modules, it will explicitly collect multi-step evidence before deriving the answer.
\end{itemize}

In both modes, the model outputs an answer $\hat{A}$ along with a confidence score $c$, which serves as a critical quality indicator for the refinement stage.

\subsection{Self-Refinement}
\label{sec:refine_stage}

To improve robustness against the rigidity of static visual programs that are produced in a single pass, \ourmethod incorporates a self-refinement mechanism designed to correct invalid executions and revise low-confidence reasoning programs:

\begin{itemize}[leftmargin=*,noitemsep]
    \item \textbf{Refinement for failed executions:}  
    When a visual program encounters a runtime failure (e.g., empty retrieval results or invalid arguments), the model inspects the execution log to diagnose the issue and refine a corrected program.

    \item \textbf{Refinement for Low-Confidence Reasoning:} 
    Even if execution completes successfully, the prediction may have low confidence. If $c < \tau$ the model modifies the reasoning program, such as broadening the retrieval scope or adjusting invoked modules, and re-executes the refined program to produce a more reliable answer.
\end{itemize}

\subsection{Training Pipeline}
\label{sec:training}
 We denote our trained VideoLLMs to execute the framework end-to-end using a two-stage pipeline: Supervised Fine-Tuning (SFT), followed by Group Relative Policy Optimization (GRPO) to further enhance reasoning quality and efficiency. To facilitate reproducibility, we provide the detailed prompts in our tasks in Appendix~\ref{sec:appendix_prompts}.
\paragraph{Supervised Fine-Tuning.}
We construct a \emph{reason-and-refine} dataset using a teacher model, covering three execution trajectories: runtime failures, successful executions with incorrect predictions, and correct predictions. The supervision is organized into two types:
\begin{itemize}[leftmargin=*,noitemsep]
    \item \textbf{Adaptive Reasoning Supervision:} 
    To train VideoLLMs to select appropriate reasoning modes, we categorize each query into different reasoning regimes. 
    Queries that can be correctly answered by the native VideoLLM with high confidence ($c>0.75$) are supervised with \texttt{Native Reasoning} (R1); 
    Otherwise, they are supervised with \texttt{Multi-step Visual Program Reasoning} (R2), where ground-truth correct predictions of visual programs are provided.

    \item \textbf{Refinement Supervision:} 
    To enable reasoning refinement, we construct supervision signals: 
    (i) Execution Failure Refinement (R3), where failed programs and their runtime logs are paired with correct predictions of visual programs; and 
    (ii) Low-Confidence / Incorrect Reasoning Refinement (R4), where programs yielding incorrect or low-confidence predictions are revised into corrected prediction programs. We provide relevant prompts in Appendix~\ref{sec:appendix_prompts}.
\end{itemize}

\paragraph{Group Relative Policy Optimization.}
To better align generation with the desired behavior, we further optimize VideoLLMs with GRPO~\cite{guo2025deepseek}. For each query, we sample a group of outputs and maximize a composite reward:
\begin{equation}
\mathcal{R} = 0.5 \cdot r_{\text{acc}} + 0.2 \cdot r_{\text{fmt}} + 0.3 \cdot r_{\text{mode}},
\end{equation}
where $r_{\text{acc}}$ rewards answer correctness, $r_{\text{fmt}}$ penalizes invalid program syntax, and $r_{\text{mode}}$ (Mode Consistency Reward, more details in Appendix~\ref{sec:appendix_reward}.) encourages selecting the reasoning mode consistent with the labels (where native VideoLLM can solve them with high confidence).

\begin{table*}[t]
\centering
\small 
\setlength{\tabcolsep}{5pt}      
\renewcommand{\arraystretch}{1.0}

\newcommand{\res}[2]{#1\fontsize{7.5pt}{7.5pt}\selectfont\textcolor{improvecolor}{\textbf{\,$\uparrow$#2}}}
\newcommand{\na}{\textcolor{gray}{--}}

\begin{tabular}{l c c c c c c} 
\toprule
\textbf{Model} & \textbf{Frames} & \textbf{LVBench} & \textbf{VideoMME}$_{L}$ & \textbf{LongVideoBench} & \textbf{MLVU} & \textbf{Avg.} \\
\midrule

\rowcolor{headergray} \multicolumn{7}{l}{\textsc{\textbf{Closed-source Models}}} \\
GPT-4o \citep{hurst2024gpt}                & 384 & 48.9 & 72.1 & 66.7 & 54.9 & 60.7 \\
OpenAI o3 \citep{openai2025o3}      & 256 & 57.1 & 64.7 & 67.5 & \na  & \na  \\
Gemini-1.5-pro \citep{team2024gemini}      & 256 & 33.1 & 67.4 & 58.6 & \na  & \na  \\
Seed1.5VL-pro \citep{seed2025seed1_5vl}    & 32  & 46.1 & 63.3 & 63.7 & 54.9 & 57.0 \\
\midrule

\rowcolor{headergray} \multicolumn{7}{l}{\textsc{\textbf{Open-source Models}}} \\
Qwen2.5-VL-72B \citep{bai2025qwen2}        & 128 & 47.4 & 64.6 & 60.3 & 53.8 & 56.5 \\
LongVILA-7B \citep{chen2024longvila}       & 256 & \na  & 52.1 & 57.7 & 49.0 & \na  \\
VideoMind-7B \citep{liu2025videomind} &  2/FPS & 40.8 & 49.2  & \na  & \na  & \na  \\
Video-R1-7B \citep{feng2025video}           & 64 & 36.2  & 48.4 & 53.9 & \na & \na  \\
Video-XL-7B \citep{shu2025video}           & 256 & \na  & 54.9 & 50.7 & 45.5 & \na  \\
Qwen2.5-VL-7B \citep{bai2025qwen2}         & 64  & 38.3 & 50.0 & 58.6 & 48.0 & 48.7 \\
Qwen3-VL-8B~\citep{yang2025qwen3}          & 64  & 40.2 & 56.3 & 61.5 & 53.6 & 52.9 \\
\midrule

\rowcolor{headergray} \multicolumn{7}{l}{\textsc{\textbf{Agentic LLMs}}} \\
VideoAgent (GPT-4) \citep{wang2024videoagent} & \na & \na   & 46.2 & \na   & 52.2 & \na \\
VideoAgent (GPT-4) \citep{fan2024videoagent}  & \na & \na   & 48.1 & \na   & 55.4 & \na \\
VideoTree (Qwen-Plus)                         & \na & \na   & 39.3 & \na   & 51.6 & \na \\
\midrule

\rowcolor{headergray} \multicolumn{7}{l}{\textsc{\textbf{Ours (Reason + Refine)}}} \\

\rowcolor{ourbg} 
Qwen2.5-VL-7B + \ourmethod                       & 64  & 47.2 & 56.7 & 60.9 & 49.6 & \res{53.6}{4.9} \\

\rowcolor{ourbg}
Qwen3-VL-8B + \ourmethod                         & 64  & \textbf{49.7} & \textbf{68.8} & \textbf{64.5} & \textbf{55.2} & \res{\textbf{59.6}}{6.7} \\

\bottomrule
\end{tabular}

\caption{\textbf{Quantitative results on long-video benchmarks.} We report performance across four benchmarks. Notably, the trained VideoLLM within our \ourmethod framework outperforms the native VideoLLM, achieving performance gains of 4.9\% and 6.7\%, respectively.}
\label{tab:video_results_main}
\end{table*}

\section{Experiments}
\label{sec:experiments}

\paragraph{Benchmarks}
We evaluate \ourmethod on four benchmarks: 
(1) LongVideoBench~\citep{wu2024longvideobench}, using the validation set across diverse durations; 
(2) {VideoMME~\citep{fu2024video}, where we focus on the long subset ($>600$s) to assess long-range reasoning; 
(3) {LVBench~\citep{wang2024lvbench}, featuring extremely long videos (up to 2h) with complex temporal logic; and 
(4) MLVU~\citep{zhou2025mlvu}, a multi-task benchmark where we report results on the test set.

\paragraph{Implementation Details}

We employ Qwen3-VL-8B~\citep{yang2025qwen3} and Qwen2.5-VL-7B~\citep{bai2025qwen2} as the backbone VideoLLM, utilizing a strong proprietary LLM, GPT5~\citep{openai_gpt5_2025} for data synthesis (which only inputs the query and few-shot examples). For the video database, we process long-duration videos by dividing them into 10-second clips. These segments are then encoded using LanguageBind\_Video~\citep{zhu2023languagebind} and paired with subtitles extracted by FFmpeg and Whisper~\citep{radford2023robust}. The framework integrates DEVA~\citep{cheng2023tracking} and EasyOCR to support object and text grounding. 
Implemented within the MS-SWIFT framework~\citep{zhao2024swiftascalablelightweightinfrastructure}, our two-stage training consists of SFT on 5k trajectories from CG-Bench~\citep{chen2024cg} for 1 epoch, followed by Group Relative Policy Optimization (GRPO) on 10k samples for 1 epoch. During inference, the model processes up to 64 video frames, automatically triggering a refinement process if the confidence score falls below $\tau=0.75$. We set $\tau=0.75$ since it already yields strong accuracy shown in Figure~\ref{fig:confidence_stats}; while higher confidence (e.g., 0.9) is even more reliable, refining all such cases would be unnecessary. The sampling temperature is set to $0.7$ when generates visual programs.

\paragraph{Baselines}
Our comparative study involves three distinct categories of state-of-the-art models:
(i) \textit{Closed-source models}, represented by GPT-4o \citep{hurst2024gpt}, Gemini-1.5 Pro \citep{team2024gemini}, and Seed-1.5VL-Pro \citep{seed2025seed1_5vl};
(ii) \textit{Open-source models}, including Qwen2.5-VL \citep{bai2025qwen2}, LongVILA-7B \citep{chen2024longvila}, Video-XL-7B \citep{shu2025video};
(iii) \textit{Agentic frameworks}: VideoAgent \citep{wang2024videoagent} and VideoTree \citep{wang2025videotree}. All models are evaluated following their official decoding configurations to show the advantages of our framework.
\begin{table*}[t]
    \centering

    \setlength{\tabcolsep}{12pt} 
    \resizebox{1.0\linewidth}{!}{
        \begin{tabular}{ll ccc}
            \toprule
            \textbf{Dataset} & \textbf{Method} & \textbf{Avg. Acc. (\%)} $\uparrow$ & \textbf{Output Len.} $\downarrow$ & \textbf{Avg. Runtime} $\downarrow$ \\
            \midrule
            
    \multirow{4}{*}{VideoMME$_{L}$} 
     & Native VideoLLM & 56.3 & 255 & 2.1s \\
     & Multi-step Visual Program & 66.7 & 1496 & 8.2s \\
     & Adaptive Reasoning & 66.8 & 825 & 5.2s \\
     \cmidrule(lr){2-5} 
     \rowcolor{ourbg} & + Refine & \textbf{68.8} & 1326 & 7.1s \\ 
    \midrule
    
    \multirow{4}{*}{LVBench} 
     & Native VideoLLM & 40.2 & 255 & 2.3s \\
     & Multi-step Visual Program & 46.4 & 1594 & 8.2s \\
     & Adaptive Reasoning & 48.1 & 1227 & 5.9s \\
     \cmidrule(lr){2-5}
     \rowcolor{ourbg} & + Refine & \textbf{49.7} & 1435 & 7.8s \\ 
            \bottomrule
        \end{tabular} }

   \caption{Performance and Efficiency Comparison. We evaluate efficiency-accuracy trade-offs across different reasoning paradigms. \ourmethod{}achieves the best balance by combining adaptive reasoning with refinement.}
    \label{tab:paradigm_ablation}
\end{table*}

\subsection{Main Results}
\label{sec:main_results}
Table~\ref{tab:video_results_main} summarizes the quantitative results on long-form video benchmarks.
By adaptively switching between \textit{native} direct answering and \textit{program-based} reasoning, \ourmethod{} consistently improves over the corresponding native VideoLLM backbones.
The largest gains appear on benchmarks that demand long-range temporal aggregation.
In particular, on LVBench and VideoMME$_{L}$, \ourmethod{} improves Qwen3-VL by more than 10\% relative.
On LongVideoBench and MLVU, the gains remain consistent but are smaller (+3.0 and +1.6 points, respectively), which we attribute to their broader mix of short- and long-horizon questions. As native VideoLLMs are already highly capable of handling the short-video subset, the overall gain is naturally averaged down.

\subsection{Ablation Studies}
\label{sec:ablation}
%

\begin{figure}[t]
    \centering
    \includegraphics[width=1.0\linewidth]{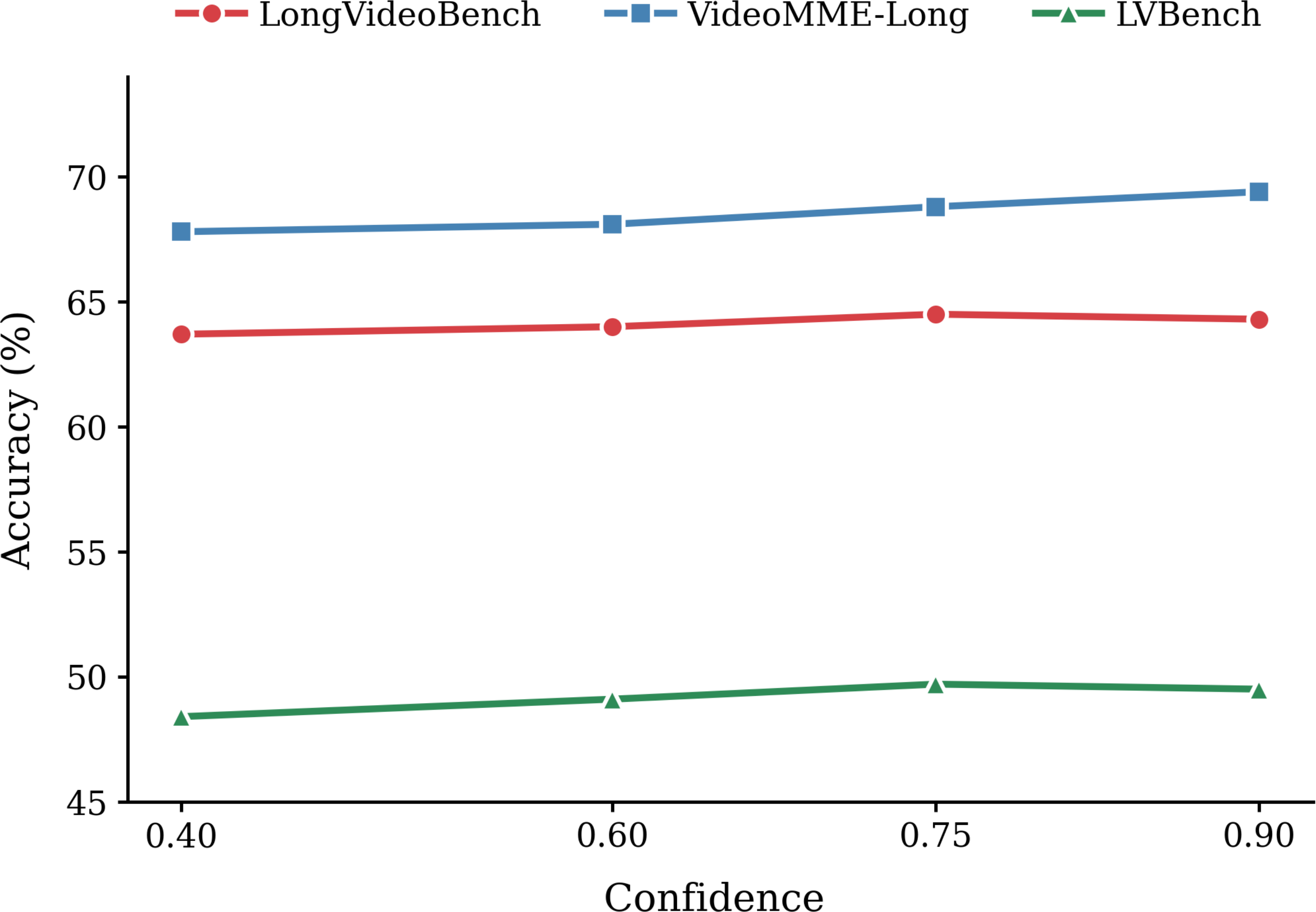}
\caption{\textbf{Performance at varying confidence thresholds.} \ourmethod exhibits robust performance on the Long Video Benchmark across the wide interval of $[0.4, 0.9]$.}
    \label{fig:conf}
\end{figure}

\paragraph{Reasoning Paradigms.}
We evaluate \ourmethod{} against three baseline paradigms: Native VideoLLM, Multi-step Visual Program, and the base Adaptive Reasoning without refinement. 
As reported in Table~\ref{tab:paradigm_ablation}, Native VideoLLM exhibits the lowest latency but suffers from inferior accuracy due to its limited reasoning depth. 
In contrast, Multi-step Visual Program improves performance at the expense of a significant increase in output length and runtime. 
The base Adaptive Reasoning achieves a more favorable trade-off by reducing computational overhead while maintaining high accuracy. 
By incorporating the refinement stage, \ourmethod{}achieves the best performance, reaching $68.8\%$ on VideoMME$_{L}$ and $49.7\%$ on LVBench. These results demonstrate that the self-refinement mechanism effectively rectifies execution failures and low-confidence reasoning programs, striking an optimal balance between reasoning quality and execution efficiency.
\paragraph{Confidence Threshold}
We examine the effect of the confidence threshold $\tau$, which serves as the decision boundary for our model's reasoning process. As shown in Figure~\ref{fig:conf}, increasing $\tau$ from 0.40 to 0.75 consistently improves performance across all benchmarks. Specifically, LVBench and LongVideoBench achieve their peak accuracy of 49.7\% and 64.5\%, respectively at $\tau=0.75$. However, further raising the threshold to 0.90 leads to a slight performance degradation on these datasets, implying that an excessively strict threshold might filter out valid cues or reasoning paths. 

\begin{table}[t]
\centering
\small 
\setlength{\tabcolsep}{4pt}
\renewcommand{\arraystretch}{1.1}
\begin{tabular}{l cc c cc}
\toprule
\multirow{2}{*}{\textbf{Method (w/o)}} & \multicolumn{2}{c}{\textbf{VideoMME$_{L}$}} & & \multicolumn{2}{c}{\textbf{LVBench}} \\
\cmidrule{2-3} \cmidrule{5-6}
& \textbf{Acc.} & \textbf{$\Delta$} & & \textbf{Acc.} & \textbf{$\Delta$} \\
\midrule
\textbf{Full Model} & \textbf{61.2} &  & & \textbf{44.5} &  \\
\midrule
\textit{w/o} Retrieval & 57.3 & \small{\color{red}{$-3.9$}} & & 42.8 & \small{\color{red}{$-1.7$}} \\
\textit{w/o} Temp. Loc. & 57.5 & \small{\color{red}{$-3.7$}} & & 42.8 & \small{\color{red}{$-1.7$}} \\
\textit{w/o} Fine-grained Vis. & 58.1 & \small{\color{red}{$-3.1$}} & & 43.9 & \small{\color{red}{$-0.6$}} \\
\textit{w/o} Global Context & 56.6 & \small{\color{red}{$-4.6$}} & & 43.8 & \small{\color{red}{$-0.7$}} \\
\bottomrule
\end{tabular}
\caption{Ablation studies on different modules using Qwen3-VL (SFT-only). We report the accuracy (\%) and the performance drop ($\Delta$). }

\label{tab:module_ablation}
\end{table}

\begin{table}[t]
\centering
\resizebox{\columnwidth}{!}{
\begin{tabular}{ll cccc}
\toprule
\multirow{2}{*}{\textbf{Base Model}} & \multirow{2}{*}{\textbf{Strategy}} & \multicolumn{2}{c}{\textbf{VideoMME$_{L}$}} & \multicolumn{2}{c}{\textbf{LVBench}} \\
\cmidrule(lr){3-4} \cmidrule(lr){5-6}
& & \textbf{Reasoning} & \textbf{+Refine} & \textbf{Reasoning} & \textbf{+Refine} \\
\midrule
\multirow{2}{*}{Qwen3-VL} 
& SFT & 61.2 & 65.0 & 44.5 & 49.1 \\
& SFT+GRPO & 66.8 & \textbf{68.8} & 48.1 & \textbf{49.7} \\
\midrule
\multirow{2}{*}{Qwen2.5-VL} 
& SFT & 52.2 & 58.1 & 42.1 & 46.7 \\
& SFT+GRPO & 53.2 & \textbf{56.7} & 45.4 & \textbf{47.2} \\
\bottomrule
\end{tabular}
}
\caption{Ablation of training strategies and refinement mechanisms across models.}
\label{tab:training_ablation}
\end{table}
\begin{figure}[t]
    \centering
    \includegraphics[width=1.0\linewidth]{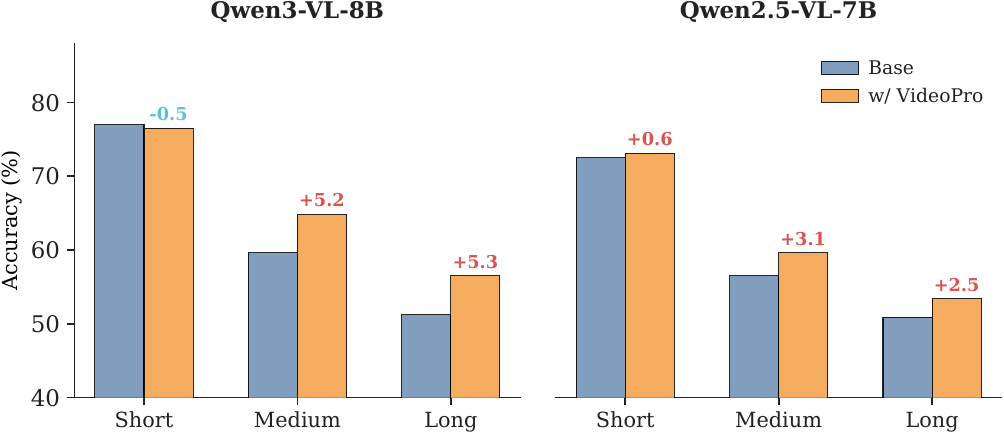}
\caption{Accuracy on LongVideoBench and VideoMME across different video durations.}
    \label{fig:duration_analysis}
\end{figure}

\begin{figure*}[t]
\centering
\includegraphics[width=1.0\linewidth]{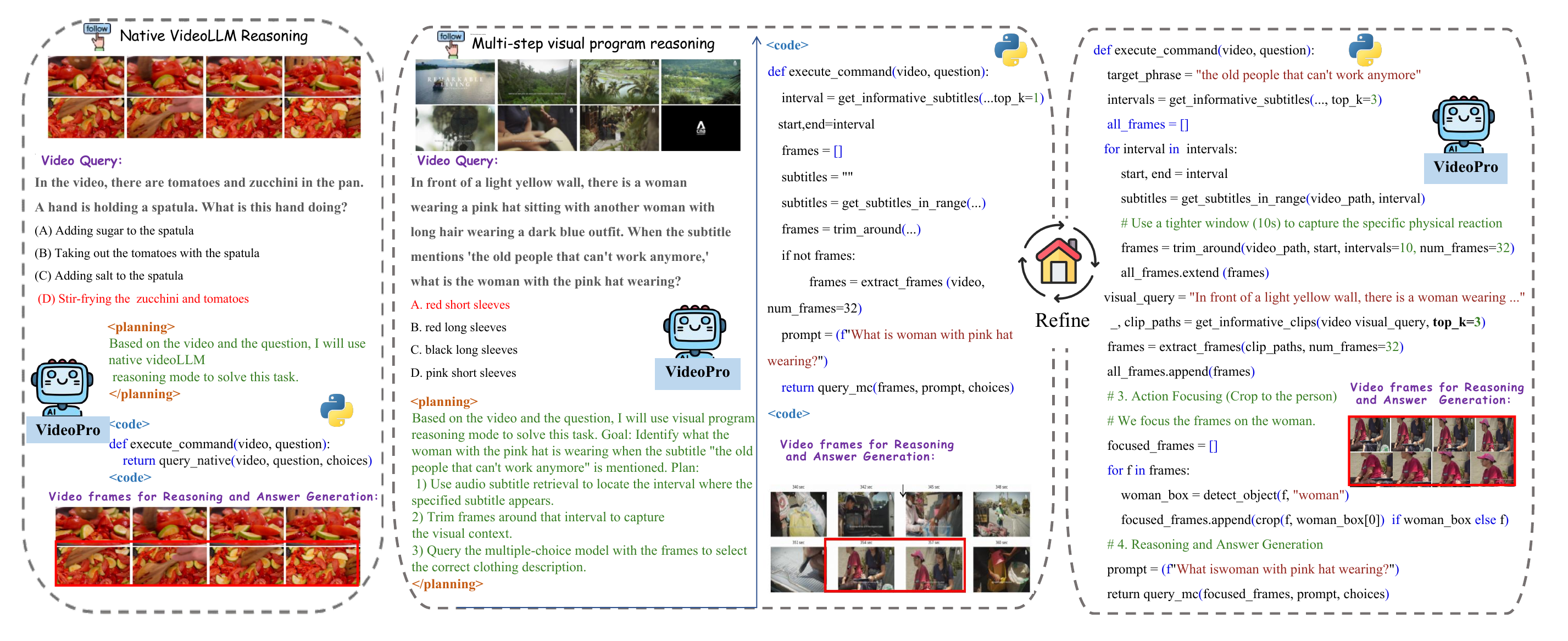}
\caption{Cases of adaptive program reasoning: native VideoLLM reasoning, multi-step visual program reasoning, and reasoning with refinement where it explicitly invokes vision modules to capture fine-grained details.}


\label{fig:case1}
\end{figure*}

\paragraph{Video Modules.}

Table~\ref{tab:module_ablation} presents an ablation study on the Qwen3-VL (SFT) backbone to evaluate the contribution of individual components within our library $\mathcal{M}$. Global Context Summarization is indispensable for long-form reasoning, as its removal causes the most significant performance drop on VideoMME$_{L}$, where the accuracy decreases from 61.2\% to 56.6\%. Regarding precise grounding, omitting either Multimodal Retrieval or Temporal Localization leads to a marked regression in LVBench accuracy, dropping to 42.8\%. Furthermore, the absence of Fine-grained Visual Extraction consistently impairs performance across all benchmarks. Collectively, these results empirically validate the necessity of our coarse-to-fine design, as the full pipeline achieves the optimal balance between efficiency and accuracy.

\paragraph{Training Strategies.}
We study the effectiveness of our two-stage learning framework by comparing models trained with SFT against our full SFT+GRPO pipeline. As shown in Table~\ref{tab:training_ablation}, while SFT provides the basic reasoning and tool-use capabilities, adding GRPO leads to a clear performance boost. For example, Qwen3-VL with base reasoning improves from 61.2\% to 66.8\% after GRPO training. This improvement shows that reinforcement learning helps the model better handle complex queries through reward alignment. Furthermore, the Refinement module consistently improves results under both training settings. For instance, it increases the SFT-only score of Qwen3-VL from 61.2\% to 65.0\%. These results prove that encouraging the model to self-correct its reasoning paths effectively reduces errors in difficult cases. Overall, the combination of GRPO training and the Refinement module achieves the best performance.

\paragraph{Video Durations.}
We evaluate \ourmethod{} across three duration groups: Short ($<2$~min), Medium ($2$--$15$~min), and Long ($>15$~min). 
As shown in Figure~\ref{fig:duration_analysis}, while performance on Short videos remains comparable to the baselines, \ourmethod{} significantly improves accuracy as duration increases. 
Specifically, for Qwen3-VL-8B, we observe absolute gains of 5.2\% and 5.3\% on the Medium and Long splits, respectively. 
This trend, where the performance gap widens with video length, shows VideoPro's efficacy in handling long videos.

\subsection{Case Study}
\label{sec:case_study}

We present representative cases in Figure~\ref{fig:case1} to show the adaptive reasoning and refinement process of VideoPro. 
For simple actions, the model utilizes \textbf{Native VideoLLM reasoning} to directly predict the action, ensuring efficiency. 
For complex long-video queries (middle), the planner invokes \textbf{Multi-step visual program reasoning} to ground specific subtitles and locate relevant intervals. 
To further enhance precision, \ourmethod performs \textbf{self-refinement}  by employing \texttt{detect\_object} and \texttt{crop} to isolate the target and filter out noise and retrieve relevant clips using scene-level textual descriptions.
Crucially, \ourmethod integrates the native ability to reason over uniformly sampled frames with the program's capability to provide localized visual evidence.

\section{Conclusion}
We propose VideoPro, a unified and adaptive reason-refine framework for long-form video understanding that explicitly balances accuracy and efficiency.
To overcome the rigidity of static inference, \ourmethod{} performs adaptive planning to select between \emph{native} mode (direct answering with the native VideoLLM) and \emph{program} mode (multi-step, executable visual programs) based on the query's reasoning demands.
To improve robustness on challenging grounding and temporal reasoning, we introduce a dual-signal self-refinement mechanism that uses execution feedback and confidence-aware triggers to detect and correct failures.
Finally, we integrate these components into a GRPO-based training pipeline, enabling \ourmethod{} to achieve strong empirical performance while producing multi-step reasoning steps through executable programs.

\section*{Limitations}

\ourmethod works well for long-form video understanding but has some limitations: (1) its temporal grounding depends on a manually curated vision-module library, reducing zero-shot flexibility and requiring manual integration or task-specific training for new categories/domains; (2) it can be overconfident (e.g., confidence $>0.9$ on wrong reasoning/answers), so self-refinement may fail when the VideoLLM is confidently wrong, potentially misleading users.

\section*{Ethics and Potential Risks}
Deploying \ourmethod raises ethical concerns about reliability in high-stakes settings: its overconfidence can cause ``silent failures'' (e.g., forensics or autonomous monitoring) where wrong outputs with high certainty bias human decisions. Because it uses LLMs for program synthesis and summarization, it is also vulnerable to hallucinations. Finally, since training data is teacher-synthesized and optimized via GRPO, societal or cultural biases in teacher models may be absorbed and amplified.

\bibliography{custom}

\clearpage
\appendix
\section{Appendix}
\label{sec:appendix}

\begin{figure*}[htbp]
\begin{lstlisting}[language=Python, caption={Some API Reference for Video Module Library}, label={lst:video_api}]
# --- 1. Multimodal Retrieval ---
def get_informative_clips(video_path, query, top_k=3, total_duration=None):
    """Retrieves visual intervals based on semantic text descriptions."""
    pass

def get_informative_subtitles(video_path, query, top_k=1, total_duration=None):
    """Retrieves intervals using audio subtitle text matching."""
    pass

# --- 2. Temporal Localization ---
def trim_around(video_path, timestamp, intervals=30, num_frames=64):
    """Extracts frames centered at a specific timestamp."""
    pass

def trim_frames(video_path, start, end, num_frames=64):
    """Retrieves frames within a specified [start, end] interval."""
    pass

# --- 3. Fine-grained Visual Extraction ---
def detect_object(frame, text, box_threshold=0.5, text_threshold=0.25):
    """Locates specific objects within a frame."""
    pass

def run_ocr(frame):
    """Extracts visual text (OCR) from the frame."""
    pass

# --- 4. Global Context Summarization ---
def get_subtitle_hints(video_path, question, choices, duration, word_number=300):
    """Summarizes narrative context from the full transcript."""
    pass

# --- 5. Reasoning and Answer Generation ---
def query_native(video_path, question, choices):
    """
    Native VideoLLM reasoning.
    Returns: (prediction, confidence_score)
    """
    pass

def query_mc(frames, question, choices):
    """
    Multiple-choice QA based on localized frames.
    Returns: (prediction, confidence_score)
    """
    pass
\end{lstlisting}

\end{figure*}

\subsection{Vision modules}
\label{appendix:modules}
We implement a general video module library to facilitate coarse-to-fine video reasoning. This library streamlines the transition from global context retrieval to fine-grained visual probing. We summarize the API reference, where each module is specialized for a distinct sub-task to ensure efficient and accurate program execution:
\begin{itemize}
    \setlength{\itemsep}{0.1em}
    \setlength{\parskip}{0pt}

    \item \textbf{Multimodal Retrieval.}  
    Retrieves semantically relevant clips and transcripts with offline embeddings efficiently, avoiding exhaustive frame-by-frame scanning and significantly reducing the search space from the frame level to the clip level.

    \item \textbf{Temporal Localization.}  
  Taking the identified temporal intervals as input, the module retrieves the corresponding video frames and subtitles. This filtering reduces irrelevant context and focuses the model on query-relevant visual and textual evidence.

    \item \textbf{Fine-grained Visual Extraction.}  
    Conducts detailed analysis on selected frames, such as object detection and optical character recognition (OCR), to extract fine visual details.

    \item \textbf{Global Context Summarization.}  
    Aggregates key thematic and narrative information from the full transcript, providing essential context for queries requiring understanding of the video’s overall storyline.

    \item \textbf{Reasoning and Answer Generation.}  
    Integrates gathered evidence to generate final answers. VideoPro is prompted to answer the question with frames by outputting a single uppercase letter.

\end{itemize}

\subsection{Prompt Details}
\label{sec:appendix_prompts}
We prompt GPT-5 with the vision-module specifications (Fig.~\ref{fig:vision_modules}) and few-shot examples (Fig.~\ref{fig:few_shot_example}) to synthesize a multi-step visual program. Concurrently, the native VideoLLM processes the query, yielding a prediction with a confidence score via \texttt{query\_native}; high-confidence outputs serve as labels for the native reasoning mode. VideoPro initiates with Adaptive Reasoning (Fig.~\ref{fig:adaptive_reasoning}), where native outputs adhere to the format in Fig.~\ref{fig:reasoning_mode1_output} and multi-step outputs follow Fig.~\ref{fig:few_shot_example}. Subsequently, VideoPro applies a refinement mechanism when program execution fails or confidence is insufficient: execution failures trigger \texttt{prompt\_refine1} (Fig.~\ref{fig:refinement_failed_program}), while low-confidence outcomes in native and multi-step reasoning utilize Fig.~\ref{fig:refinement_prompt} and Fig.~\ref{fig:refinement_low_confidence_program}, respectively.

\subsection{Mode Consistency Reward}
\label{sec:appendix_reward}
We design the mode consistency reward to encourage the model to select the native reasoning mode only when the backbone VideoLLM can solve the query correctly and confidently without invoking auxiliary modules.
For each training sample, we first run \texttt{query\_native} to obtain a native prediction and its confidence score \((\hat{A}_{\mathrm{nat}}, c_{\mathrm{nat}})\), where \(c_{\mathrm{nat}}\) is computed from the output token probabilities.
We then define an oracle mode label \(m^\star\) as:
\[
m^\star =
\begin{cases}
\mathrm{native}, & \text{if } \hat{A}_{\mathrm{nat}} = y \text{ and } c_{\mathrm{nat}} \ge \tau,\\
\mathrm{program}, & \text{otherwise},
\end{cases}
\]
where \(y\) is the ground-truth answer and \(\tau\) is a confidence threshold (0.75).
Finally, we define
\[
r_{\mathrm{mode}} = \mathbb{I}[m = m^\star],
\]
So the model is rewarded for selecting the native reasoning mode on simple queries and the program reasoning mode otherwise.

\begin{table}[t]
    \centering
    \footnotesize
    \setlength{\tabcolsep}{2pt} 
    \renewcommand{\arraystretch}{0.9} 
    \begin{tabular}{lccc}
        \toprule
        Dataset & Native (\%) & Program (\%) & Refinement (\%) \\
        \midrule
        LongVideoBench & 29.4 & 70.6 & 35.9 \\
        LVBench        & 17.7 & 82.3 & 55.5 \\
        \bottomrule
    \end{tabular}
    \caption{Distribution of execution modes and refinement ratios.}
    \label{tab:mode_refine_stats}
\end{table}
\subsection{Execution Mode Analysis}
Table \ref{tab:mode_refine_stats} summarizes the distribution of execution modes in VideoPro. The majority of queries across both benchmarks are processed via multi-step program reasoning. This is particularly evident on LVBench, where 82.3\% of tasks utilize the program mode. Such results highlight the necessity of multi-step reasoning for complex long-video understanding. Additionally, LVBench triggers the refinement mechanism more frequently than LongVideoBench, reaching a ratio of 55.5\%.

\begin{figure*}[t]
\centering
\small 
\begin{AIbox}{Vision Modules API Definitions}{
    \textbf{1. Multimodal Retrieval} (Offline semantic search)
    \begin{itemize}
        \item \texttt{get\_informative\_clips(video, query, k=3)}: Returns relevant visual intervals.
        \item \texttt{get\_informative\_subtitles(video, query, k=1)}: Returns intervals based on audio text.
    \end{itemize}

    \textbf{2. Temporal Localization} (Context filtering)
    \begin{itemize}
        \item \texttt{trim\_\{after, before, around, frames\}(...)}: Slices video into target temporal segments.
    \end{itemize}

    \textbf{3. Fine-grained Visual Extraction} (Local detail analysis)
    \begin{itemize}
        \item \texttt{detect\_object(frame, text)}: Returns bounding boxes for specified objects.
        \item \texttt{run\_ocr(frame)}: Extracts visual text; \texttt{crop(frame, box)}: Zooms into regions.
    \end{itemize}

    \textbf{4. Global Context Summarization} (Narrative aggregation)
    \begin{itemize}
        \item \texttt{get\_subtitle\_hints(video, question, choices)}: Summarizes full-transcript evidence.
    \end{itemize}

    \textbf{5. Reasoning and Answer Generation} (VideoLLM inference)
    \begin{itemize}
        \item \texttt{query\_native(video, q, c)}: Native VideoLLM reasoning for direct option selection.
        \item \texttt{query\_mc(frames, q, c)} / \texttt{query\_yn(frames, q)}: VQA over localized frames.
    \end{itemize}
}
\end{AIbox}
\caption{The predefined API library for our vision modules. The system decomposes complex queries into sub-tasks using these five functional blocks to bridge the gap between raw video data and program reasoning.}
\label{fig:vision_modules}
\end{figure*}
\begin{figure*}[t]
\begin{AIbox}{Few-shot Example for Visual Program Reasoning}{

\textbf{Question:} In the video, a woman wearing glasses and a brown top folded the lower right corner of the book. What vegetables were on the book?

\textbf{Choices:}
\begin{itemize}
    \item A. pumpkins \quad B. beans \quad C. corn \quad D. carrots \quad E. squash \quad F. peas \quad G. potatoes
\end{itemize}

\vspace{1em}
\textless planning\textgreater

Based on the video and the question, I will use the visual program reasoning mode to solve this task. First, retrieve clips showing the woman in glasses and a brown top, folding the book. Then, filter frames containing the book. If no such frames are found, use all frames. Finally, query the model to identify the vegetables on the book from the given choices.

\textless /planning\textgreater

\vspace{1em}
\textless code\textgreater

\texttt{def execute\_command(video, question):}\\
\hspace*{2em}\texttt{intervals, clip\_paths = get\_informative\_clips(}\\
\hspace*{4em}\texttt{video,}\\
\hspace*{4em}\texttt{"woman wearing glasses and a brown top folding the book",}\\
\hspace*{4em}\texttt{top\_k=1,}\\
\hspace*{2em}\texttt{)}\\
\hspace*{2em}\texttt{frames = []}\\
\hspace*{2em}\texttt{for clip in clip\_paths:}\\
\hspace*{4em}\texttt{frames.extend(extract\_frames(clip))}\\
\hspace*{2em}\texttt{book\_frames = []}\\
\hspace*{2em}\texttt{for frame in frames:}\\
\hspace*{4em}\texttt{if detect\_object(frame, "book"):}\\
\hspace*{6em}\texttt{book\_frames.append(frame)}\\
\hspace*{2em}\texttt{if not book\_frames:}\\
\hspace*{4em}\texttt{book\_frames = frames}\\
\hspace*{2em}\texttt{prompt = "What vegetables were on the book?"}\\
\hspace*{2em}\texttt{return query\_mc(book\_frames, prompt, choices)}

\textless /code\textgreater
}
\end{AIbox}
\caption{\textbf{Few-shot Example of Visual Program reasoning.}}
\label{fig:few_shot_example}
\end{figure*}

\begin{figure*}[t]
\begin{AIbox}{Adaptive Reasoning Prompt}{

You will receive a multiple-choice question about a video. Your output must define a Python function in the following format:

\vspace{1em}
\textless planning\textgreater

Briefly judge whether the question can be direct solved by videoLLM and plan the main API calls and reasoning steps you will use.
\begin{itemize}
    \item If it can be answered, use native reasoning mode with a single \texttt{query\_native} call.
    \item If it needs long-range or multi-step reasoning, use more detailed visual program mode with other modules (retrieval, subtitles, frame analysis, etc.).
\end{itemize}
\textless /planning\textgreater

\vspace{1em}
\textless code\textgreater

Write a Python function in the following format.

\vspace{0.5em}
\texttt{def execute\_command(video, question):}\\
\hspace*{2em}\texttt{\# Visual program code (no comments needed inside the code body).}\\
\hspace*{2em}\texttt{...}\\
\hspace*{2em}\texttt{return answer}

\textless /code\textgreater
}
\end{AIbox}
\caption{\textbf{Adaptive Reasoning Prompt Template. (Input for R1 \& R2)}}
\label{fig:adaptive_reasoning}
\end{figure*}

\begin{figure*}[t]
\begin{AIbox}{Native VideoLLM Reasoning Output}{

\textless planning\textgreater

Based on the video and the question, I will use native videoLLM reasoning mode to solve this task.

\textless /planning\textgreater

\vspace{1em}
\textless code\textgreater

\texttt{def execute\_command(video, question):}\\
\hspace*{2em}\texttt{return query\_native(video, question, choices)}

\textless /code\textgreater
}
\end{AIbox}
\caption{\textbf{Example Output for Native VideoLLM Reasoning Mode. (Output for R1)}}
\label{fig:reasoning_mode1_output}
\end{figure*}

\begin{figure*}[t]
\begin{AIbox}{Refinement for Failed Program}{

You will receive a multiple-choice question about a video and a Python visual program in the \texttt{execute\_command} format, and a runtime error log from running this program.

\vspace{1em}
\texttt{\{question\_with\_choices\}}

\vspace{1em}
Buggy visual program:
\vspace{0.5em}
\{\}

\vspace{1em}
Runtime error log:
\vspace{0.5em}
\{\}

\vspace{1em}
Refine this visual program by fixing the bugs.
}
\end{AIbox}
\caption{\textbf{Refinement Prompt for Failed Program. (Input for R3)}}
\label{fig:refinement_failed_program}
\end{figure*}

\begin{figure*}[t]
\begin{AIbox}{Refinement for Low-Confidence Native Reasoning}{

You will receive a multiple-choice question about a video and an existing visual program that only uses the native-mode helper API \texttt{query\_native}.

\vspace{1em}
\texttt{\{question\_with\_choices\}}

\vspace{1em}
Current native visual program:

\textless code\textgreater

\texttt{def execute\_command(video, question):}\\
\hspace*{2em}\texttt{return query\_native(video, question, choices)}

\textless /code\textgreater

\vspace{1em}
Refine this visual program.
}
\end{AIbox}
\caption{\textbf{Refinement Prompt for Low-Confidence Native Reasoning. (Input for R4)}}
\label{fig:refinement_prompt}
\end{figure*}

\begin{figure*}[t]
\begin{AIbox}{Refinement for Low-Confidence Program Reasoning}{

You will receive a multiple-choice question about a video and an existing visual program.

\vspace{1em}
\texttt{\{question\_with\_choices\}}

\vspace{1em}
Current visual program:
\vspace{0.5em}
\{\}

\vspace{1em}
Refine this visual program to improve its reasoning and correctness.
}
\end{AIbox}
\caption{\textbf{Refinement Prompt for Low-Confidence Program Reasoning. (Input for R4)}}
\label{fig:refinement_low_confidence_program}
\end{figure*}

\begin{figure*}[t]
\begin{AIbox}{Prompt for Answer Generation}{

Select the best answer to the following multiple-choice question based on the video. Respond with only the letter (A, B, C, or D or other letter) of the correct option.

\vspace{1em}
\texttt{\{question\_with\_choices\}}

\vspace{1em}
Output a single letter. The best answer is:
}
\end{AIbox}
\caption{Prompt for Answer Generation. This prompt guides the model to output a distinct uppercase letter prediction consistent with the vision module interface.}
\label{fig:native_answer_prompt}
\end{figure*}

\end{document}